%% file: main.tex
\documentclass[10pt,twocolumn,letterpaper]{article}

\usepackage[pagenumbers]{cvpr} % To force page numbers, e.g. for an arXiv version

\usepackage{graphicx}
\usepackage{amsmath}
\usepackage{amssymb}
\usepackage{booktabs}
\usepackage{multirow}
\usepackage{siunitx}
\usepackage{bbding}

\usepackage[pagebackref=true,breaklinks=true,letterpaper=true,colorlinks,bookmarks=false]{hyperref}
\usepackage{makecell}
\usepackage{xcolor} %% for textcolor gray in supply
\usepackage{pifont}

\newcommand{\eat}[1]{}

\makeatletter 
  \newcommand\figcaption{\def\@captype{figure}\caption} 
  \newcommand\tabcaption{\def\@captype{table}\caption} 
\makeatother

\usepackage[capitalize]{cleveref}
\crefname{section}{Sec.}{Secs.}
\Crefname{section}{Section}{Sections}
\Crefname{table}{Table}{Tables}
\crefname{table}{Tab.}{Tabs.}

\def\cvprPaperID{11810} % *** Enter the CVPR Paper ID here
\def\confName{CVPR}
\def\confYear{2023}

\usepackage{caption} 
\begin{document}


% \title{Generative Visual Dialog for Interactive Image Understanding}
%\title{InformDialog: Interaction between Vision and Language Foundations}

% \title{InfoVisDial: Generating Informative Visual Dialog by Bridging Multimodal and Language Foundation Modele}

\title{
InfoVisDial: An Informative Visual Dialogue Dataset \\
by Bridging Large Multimodal and Language Models
}

\author{
Bingbing Wen$^{1}$\thanks{Work done while interning at Microsoft},
Zhengyuan Yang$^{2}$, 
Jianfeng Wang$^2$,
Zhe Gan$^2$, 
Bill Howe$^1$,
Lijuan Wang$^2$
\\
$^1$University of Washington, USA
\qquad
$^2$Microsoft Azure AI, USA
\\
}
\maketitle

%%%%%%%%% ABSTRACT
\begin{abstract}
  \input{abs2}
\end{abstract}

\input{intro}

\input{related}

\input{approach}

\input{dataset}

\input{exp}
\input{conclusion}

%%%%%%%%%%%%%%%%%%%%%%%%%%%%%%%%%%%%%%%%
\clearpage
{\small
\bibliographystyle{ieee_fullname}
\bibliography{egbib}
}

% %%%%%%%%%%%%%%%%%%%%%%%%%%%%%%%%%%%%%%%%
% \clearpage
% \appendix
% \input{supply}
% %%%%%%%%%%%%%%%%%%%%%%%%%%%%%%%%%%%%%%%%

\end{document}

%% file: abs2.tex
In this paper, we build a visual dialogue dataset, named InfoVisDial, which provides rich informative answers in each round even with external knowledge related to the visual content. Different from existing datasets where the answer is compact and short, InfoVisDial contains long free-form answers with rich information in each round of dialogue. For effective data collection, the key idea is to bridge the large-scale 
multimodal model (e.g., GIT) and the language models (e.g., GPT-3). GIT can describe the image content even with scene text, while GPT-3 can generate informative dialogue based on the image description and appropriate prompting techniques. With such automatic pipeline, we can readily generate informative visual dialogue data at scale. Then, we ask human annotators to rate the generated dialogues to filter the low-quality conversations.
Human analyses show that InfoVisDial covers informative and diverse dialogue topics: $54.4\%$ of the dialogue rounds are related to image scene texts, and $36.7\%$ require external knowledge. Each round's answer is also long and open-ended: $87.3\%$ of answers are unique with an average length of $8.9$, compared with $27.37\%$ and $2.9$ in VisDial.
Last, we propose a strong baseline by adapting 
the GIT model for the visual dialogue task and fine-tune the model on InfoVisDial. Hopefully, our work can motivate more effort on this direction.

%% file: intro.tex
\section{Introduction}

\begin{figure}[t]
\begin{center}
\centerline{\includegraphics[width=0.95\linewidth]{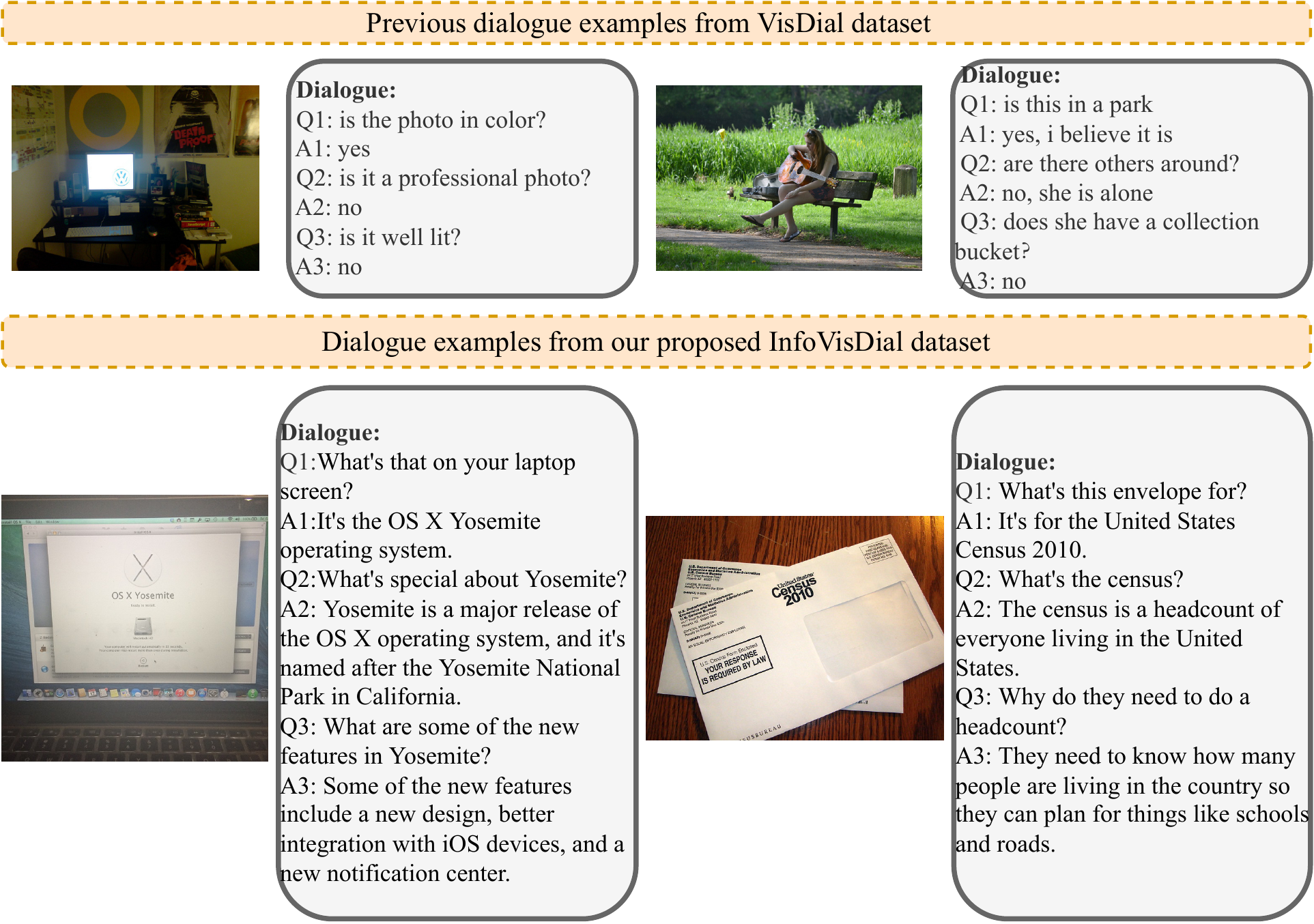}}
\end{center}
\vspace{-3mm}
\caption{
% Compared with dialogues from VisDial dataset whose answer is compact, dialogues from our newly proposed dataset InfoVisDial provide long free-form answers with rich information.
Examples of VisDial~\cite{das2017visual} and our InfoVisDial datasets.
Answers in VisDial~\cite{das2017visual} are typically compact and short, while those in our InfoVisDial are rich, informative, and knowledgable.
}
\label{fig:intro}
\end{figure}

\begin{table*}[t]
\centering
\scalebox{0.9}{
\begin{tabular}{cccccccccc}
\toprule
Dataset & Multi-turn& Visual Component & Scene Text& Knowledge & Long Answer \\ \midrule
VisDial~\cite{das2017visual} &\checkmark &\checkmark & &   &\\
QuAC~\cite{choi-etal-2018-quac} &\checkmark & &&\checkmark  & \checkmark \\
CoQA~\cite{reddy2019coqa} &\checkmark & & &\checkmark &\checkmark \\
TextVQA~\cite{singh2019towards} & & \checkmark&\checkmark & & \\
OKVQA~\cite{marino2019ok} & & \checkmark& &\checkmark & \\
\midrule
InfoVisDial (Ours) &\checkmark & \checkmark & \checkmark  & \checkmark & \checkmark \\
\bottomrule
\end{tabular}
}
  \caption{Comparison of 
%   the newly collected 
  our
  InfoVisDial dataset to other dialogue and visual question answering datasets.}
\label{tab:dataset overview}
\end{table*}

%%%%%%%%%%%%%%%%%%%%%%%%%%%%%%%%%%%%
Visual dialogue is a vision-language (VL) problem that aims to build AI models that could interact with humans based on the understanding of input images. Such systems can have many potential applications, ranging from information-seeking chatbots~\cite{zeng2022socratic}, interactive visual descriptions, to assisting visually-impaired people.
% VisDial~\cite{das2017visual} is a popular dataset developed for this task, and it provides a common playground for researchers to develop different visual dialogue models.
VisDial~\cite{das2017visual} is a widely-used dataset developed for this task,
and numerous approaches have been proposed based on the benchmark.
For example, prior works~\cite{kottur2018visual, guo2019image} have focused on designing various attention mechanisms to capture the interactions between the question, dialogue history, and the input image. Other studies~\cite{chen2022utc,wang2020vd} have leveraged pre-trained models with new introduced losses. 
However, prior studies mainly focus on the dialogue model design but neglect the importance of the visual dialogue problem setup and the corresponding data. 
The output texts in current datasets are typically compact and short, \eg, an average answer length of $2.9$ on the VisDial dataset~\cite{das2017visual}. The short answer intuitively contains limited information nor external knowledge about the visual content. Therefore, we aim to develop a new visual dialogue dataset with more \emph{informative} dialogue responses, \eg, a descriptive answer containing the scene text
or the external information.
Figure~\ref{fig:intro} shows examples of our dataset and the comparison to VisDial.
% \ie, \zyang{informative here? scene text & knowledge?}
% The output texts are typically short and not informative due to the limitation of the training dataset. And actually, the average answer length of the VisDial dataset~\cite{das2017visual} is only 2.9. Therefore, we believe new visual dialogue datasets about more \emph{informative} dialogue responses may shed light on the scarce landscape of visual dialog tasks.

%%%%%%%%%%%%%%%%%%%%%%%%%%%%%%%%%%%%
The classic approach for dataset collection is via human annotations, \eg, with two annotators talking to each other~\cite{das2017visual}.
Although crowd-sourcing is one popular approach to create new datasets, it is expensive and could bring annotation artifacts~\cite{schwartz-etal-2017-effect}. Inspired by the 
% growing ability
strong generalization capability
of large pre-trained language models~\cite{brown2020language}, we explore automatic dataset curation methods for informative visual dialogues. Generating informative visual dialogue requires two core capabilities. First, to understand the input visual content, the model needs the multimodal understanding ability to link visual entities with the corresponding textual concepts. Second, to generate fluent text and provide relevant information during the conversation, the model also needs strong language modeling ability to understand the dialogue history and generate proper responses. 
Luckily, recent large-scale vision-language models (\emph{e.g.}, SimVLM~\cite{wang2021simvlm}, GIT~\cite{wang2022git}, and CoCa~\cite{yu2022coca}) and language models (\emph{e.g.}, GPT-3~\cite{brown2020language} and PaLM~\cite{chowdhery2022palm}) have shown such capabilities. On the one hand, vision-language models perform well in converting visual contents into textual descriptions, but alone lack the ability of open-ended dialogue. On the other hand, language models show strong finetuned or even zero-/few-shot dialogue capabilities, but alone do not understand visual contents. 
% To this end, we propose to bridge these two types of foundation models to generate visual dialogue data on the fly. 
%by synergistically combining a VL and language model, with two possible designs.

%%%%%%%%%%%%%%%%%%%%%%%%%%%%%%%%%%%%%%%%%%%%%%%%%%%%%%%%%%%%%%%%%%

%%%%%%%%%%%%%%%%%%%%%%%%%%%%%%%%%%%%%%%%%%%%%%%%%%%%%%%%%%%%%%%%%%
%%%%%%%%%%%%%%%%%%%%%%%%%%%%%%%%%%%%
%In this study, we propose \textbf{InfoVisDial}, a framework for generative visual dialogue. 
To this end, we propose to bridge these two types of foundation models to generate visual dialogue data on the fly. 
Specifically, inspired by previous studies on prompting language models for vision-language tasks~\cite{yang2022empirical,xie2022visual,zeng2022socratic,wang2022language}, we convert the input image into textual descriptions at different granularities with a pre-trained GIT model~\cite{wang2022git}, and uses GPT-3~\cite{brown2020language} to paraphrase the related contents into a piece of dialogue in a few-shot manner. We examine different prompts and textual descriptions for an effective recipe for generating visual dialogue. 
% As shown in Figure~\ref{fig:intro},
By doing so, we can already generate reasonable visual dialogues with free-form texts and open-ended topics as shown in Figure~\ref{fig:intro}.

% {\color{red}Given an image of a can, our model could first recognize that it is a kolsch beer, and generate a very informative dialogue about its originality, taste and brew style. The questions ``what does that mean? '' in the first turn and ``where does this beer come from?''  in the third turn require reading and reasoning about the scene text. The second question ``What makes this beer German Style?'' could not be answered solely based on the image, which requires the model to be knowledgeable about the kolsch beer.  Most importantly, these generated answers are impressive. Unlike previous visual dialogue~\cite{das2017visual} setting, where the answer usually stops at ``Kolsch is a type of beer'', here, the response not only answers the first question correctly, but also provides excessive information about kolsch beer. }

We collect dialogues based on images from the TextVQA and TextCap dataset~\cite{singh2019towards,sidorov2020textcaps}, which is a scene-text-rich subset selected from the OpenImages dataset. We name our dataset as InfoVisDial, as the dialogues now become more informative and human-like. As summarized in Table~\ref{tab:dataset overview}, compared with existing dialogue and visual question answering datasets, InfoVisDial is the only dataset 
% that is multi-turn and scene-text rich, contains visual component, knowledge and long answers. 
that contains multiple turns, scene text, visual components, 
knowledge and free-form long answers. 
We collect dialogues by prompting GPT-3 with ground-truth captions or captions generated by GIT, and involve human annotators to perform fact checking and filter out low-quality dialogues. 

%%%%%%%%%%%%%%%%%%%%%%%%%%%%%%%%%%%%
% Despite first showing the informative visual dialogue capability, combining VL and language models via prompting has two limitations. First, requiring both the VL and language models during inference makes the dialogue system cumbersome. Second, visual information is represented as a textual description, which could cause an information bottleneck and lose certain visual details~\cite{yang2022empirical}. To address these two limitations, instead of inference-time prompting, we propose to distill the open-ended dialogue capability from GPT-3 to GIT by finetuning the pre-trained GIT model on the collected new InfoVisDial dataset. To this end, we obtain a single model to generate informative visual dialogue with no intermediate textual descriptions. %Specifically, we finetune the generative image-to-text (GIT) model~\cite{wang2022git} with dialogues generated by GPT-3, optionally with human cleaning.
% In terms of evaluation, we use caption and word accuracy metrics~\cite{choi-etal-2018-quac} to evaluate the text quality and content accuracy, respectively. 

With the new dataset, we adapt the GIT framework by 
adding all the history dialogues and the current question as
the text prefix, and train the model to complete the text as answers.
The fine-tuned model effectively absorbs GPT-3's dialoguing capability.
As the answer is in free form, it is challenging to automatically evaluate the performance. Motivated by the captioning task, we also use the captioning metric in addition to the key-word accuracy metrics~\cite{choi-etal-2018-quac} to evaluate the model.
The metrics also align well with human evaluation.

%%%%%%%%%%%%%%%%%%%%%%%%%%%%%%%%%%%%
%In order to eliminate the constraints of GPT-3 and
%empower other VL models’ dialogue ability, we propose a new dataset also denoted as InformDialog shown in Table~\ref{tab:dataset overview} which is collected from the informative dialogue corpus. 

% Furthermore, we also experiment on the VisDial~\cite{das2017visual} dataset to evaluate the effectiveness of the proposed approach of distilling language information from the language model by finetuning the VL model. Performance.

\begin{figure*}[t]
\begin{center}
  \centerline{\includegraphics[width=0.8\linewidth]{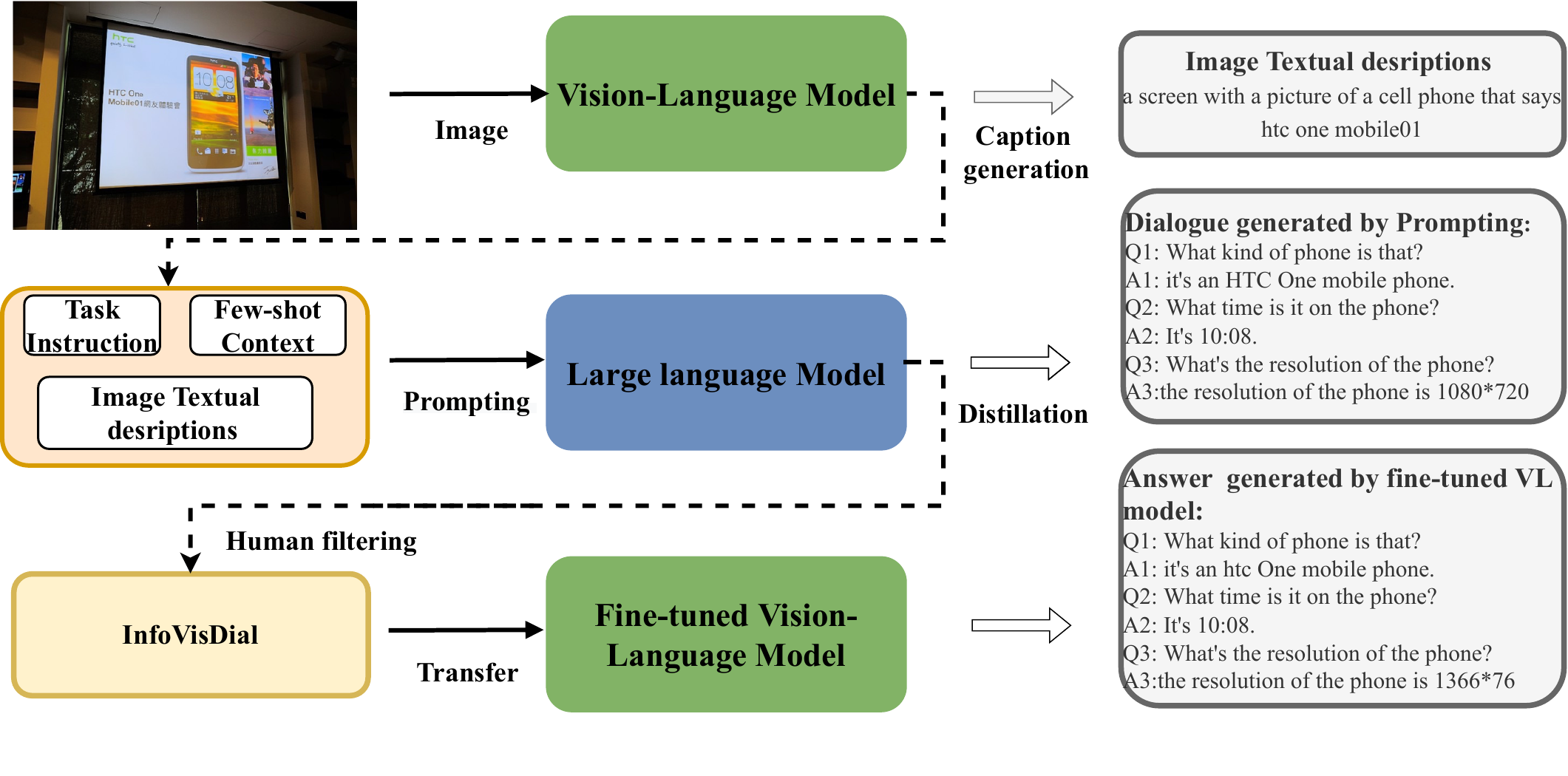}}
\end{center}
\vspace{-0.4in}
    \caption{ We curate the InfoVisDial dataset by proposing a new framework that generates informative visual dialogues by bridging multimodal (GIT) and language foundation models (GPT-3) and letting them interplay with each other. GIT converts the input image into textual descriptions. Then, we prompt GPT-3 with these textual descriptions and get the informative visual dialogue. This process could also be understood as knowledge distillation from GPT-3 to corpus. With human filtering, we could get InfoVisDial dataset. Then, we fine-tune GIT on our newly collected dataset to empower GIT with the dialogue ability.
	}
\label{fig:framework}
\end{figure*}

Our contributions are summarized as follows.
%%%%%%%%%%%%%%%%%%%%%%%%%%%%%%%%%%%%%%%%
\begin{itemize}
\item We present a new framework that generates visual dialogues by bridging multimodal and language foundation models and letting them interplay with each other, which provides a solution for automatic visual dialogue dataset curation.

\item We introduce a newly collected dataset InfoVisDial, 
%based on our proposed framework which distills knowledge from the language model into a independent corpus. The dataset 
which contains informative visual dialogues that require the model to read scene text and reason about relevant knowledge. We present a detailed analysis of the dataset, providing valuable insights on the challenges of this new task.

\item We finetune the pre-trained GIT model on the new InfoVisDial dataset, %to distil the knowledge transfering from dataset corpus.
and conduct experiments to show the effectiveness of our finetuned GIT model. We hope our new dataset can provide a solid new playground for researchers interested in the visual dialogue task.
\end{itemize}

%% file: related.tex
\section{Related Work}

\noindent\textbf{Visual dialogue.}
Compared with visual question answering~\cite{antol2015vqa, goyal2017making, anderson2018bottom, marino2019ok,hudson2019gqa}, visual dialogue not only requires that the agent is able to answer the question, but also demands the agent to fully utilize the information in previous questions and answers. Prior visual dialogue models~\cite{kottur2018visual, guo2019image,zheng2019reasoning} focus on designing various attention mechanisms to capture the interactions between a question, a dialogue history and an image. With the increasing attention paid to pre-trained models, transformer-based models with various structures are also introduced into visual dialogue tasks~\cite{murahari2020large,nguyen2020efficient}. VD-BERT~\cite{wang2020vd} leverages the pre-trained BERT model for visual dialogue task. Instead of taking discriminative and generative tasks separately, some recent work explores the relationship between these two tasks~\cite{gan2019multi,nguyen2020efficient} by training two decoders simultaneously. UTC~\cite{chen2022utc} combines discriminative and generative tasks into a unified transformer by introducing two inter-task contrastive losses.
Apart from model improvement, GST\cite{kang2022dialog} adopts semi-supervised learning and produces machine-generated visual dialogues which are similar to VisDial dataset for data augmentation. 

However, the output texts of these models are still short and not human-like due to the limitation of the training dataset (the average answer length of VisDial dataset is 2.9). Since the research area of informative visual dialogues is under-explored,  our newly proposed dataset InfoVisDial might shed light on the scarce landscape of visual dialogue.

\vspace{2mm}
\noindent\textbf{Vision-language pre-training.}
Vision-Language pre-training (VLP) has achieved rapid progress in the vision-language community aiming to learn joint multimodal representations for downstream tasks. From the representation perspective, CLIP~\cite{radford2021learning}, ALIGN~\cite{jia2021scaling}, and Florence~\cite{yuan2021florence} encode image and text into a common latent space to align representations from various modalities. They show strong zero-shot ability in image classification tasks, \emph{etc}. From the generation perspective, SimVLM~\cite{wang2021simvlm},  CoCa~\cite{yu2022coca}, GIT~\cite{wang2022git} and many others adopt encoder-decoder models trained with generative losses. These models demonstrate remarkable zero-shot/few-shot abilities in image captioning, visual question answering and \emph{etc}.  There are also other studies bridging the gap between vision-language and language models.  For example, PICa~\cite{yang2022empirical} adopts GPT-3 to extract commonsense knowledge for visual question answering.

Socratic Models~\cite{zeng2022socratic} use LM prompting to leverage multiple pre-trained models to generate assistive dialogue. However, our approach not only generates informative dialogue but also propose a new dataset curation method for low-resource area such as visual dialogue. Furthermore, by distilling the informative dialogue ability from GPT-3 to VL models, we could eliminate the constraints of GPT-3 and empower the VL models' dialogue ability. 

\vspace{2mm}
\noindent\textbf{Distilling knowledge from large language models.} Knowledge distillation draws increasing attention recently due to the rise of massive pre-trained LMs such as GPT-3~\cite{brown2020language}, which achieves huge success in downstream NLP tasks and superior in-context few-shot learning ability. Symbolic Knowledge Distillation~\cite{west2021symbolic} follows a model-corpus-model knowledge distillation strategy and  extracts the commonsense knowledge from GPT-3 into two forms: a large commonsense knowledge graph and a compact commonsense model. However, our method does not take GPT-3 as the teacher. GPT-3 and VL models are equally interacting with each other. Specifically, GIT provides GPT-3 with the textual representation of images and help GPT-3 understand visual content. 
Note that knowledge distillation usually requires a large amount of data to guarantee the transfer quality, which may not be easily obtained in real-world applications. Our method does not have the data bottleneck in nature, since we could generate as much data as we want. 

\begin{figure*}[t]
\centering
\minipage{0.45\textwidth}
  \includegraphics[width=\linewidth]{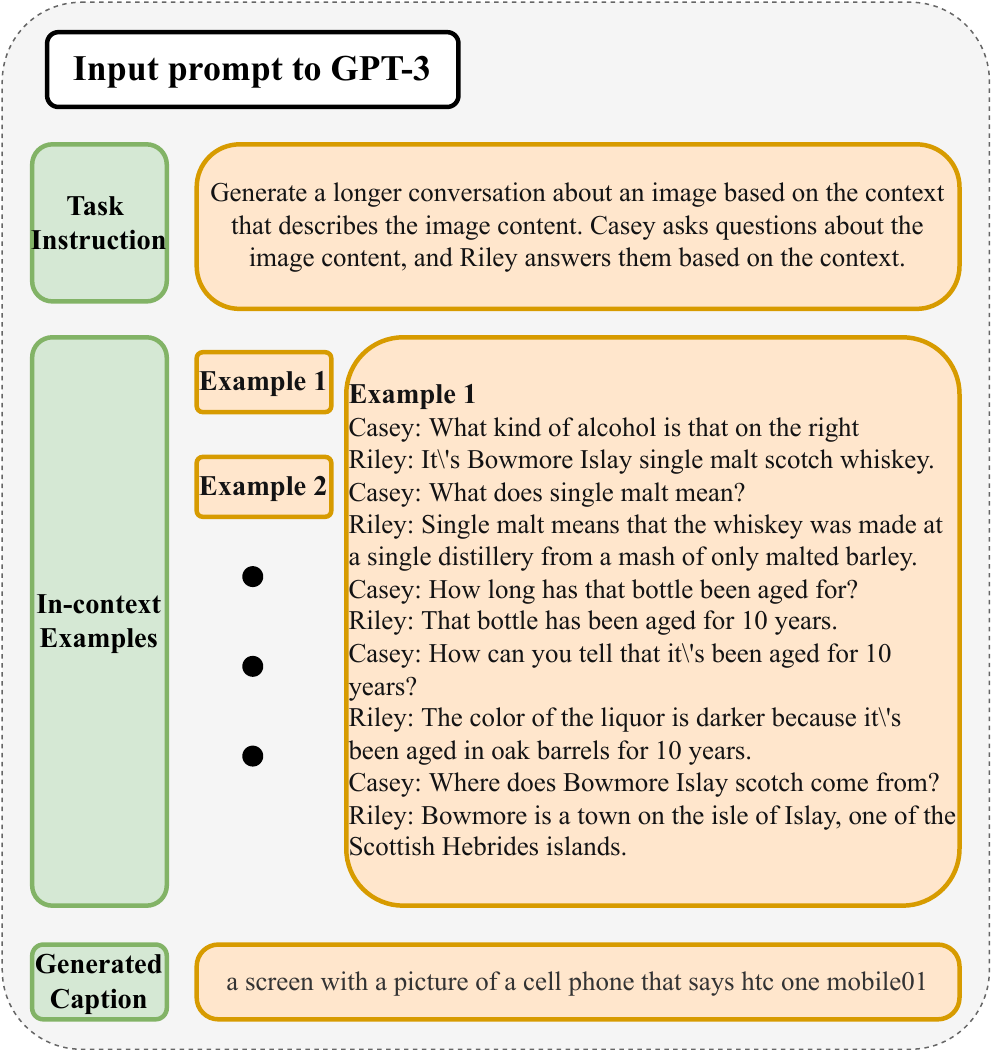}
  \caption{Example of few-shot prompt.}\label{fig:prompts}
\endminipage\hfill
\minipage{0.45\textwidth}
  \includegraphics[width=\linewidth]{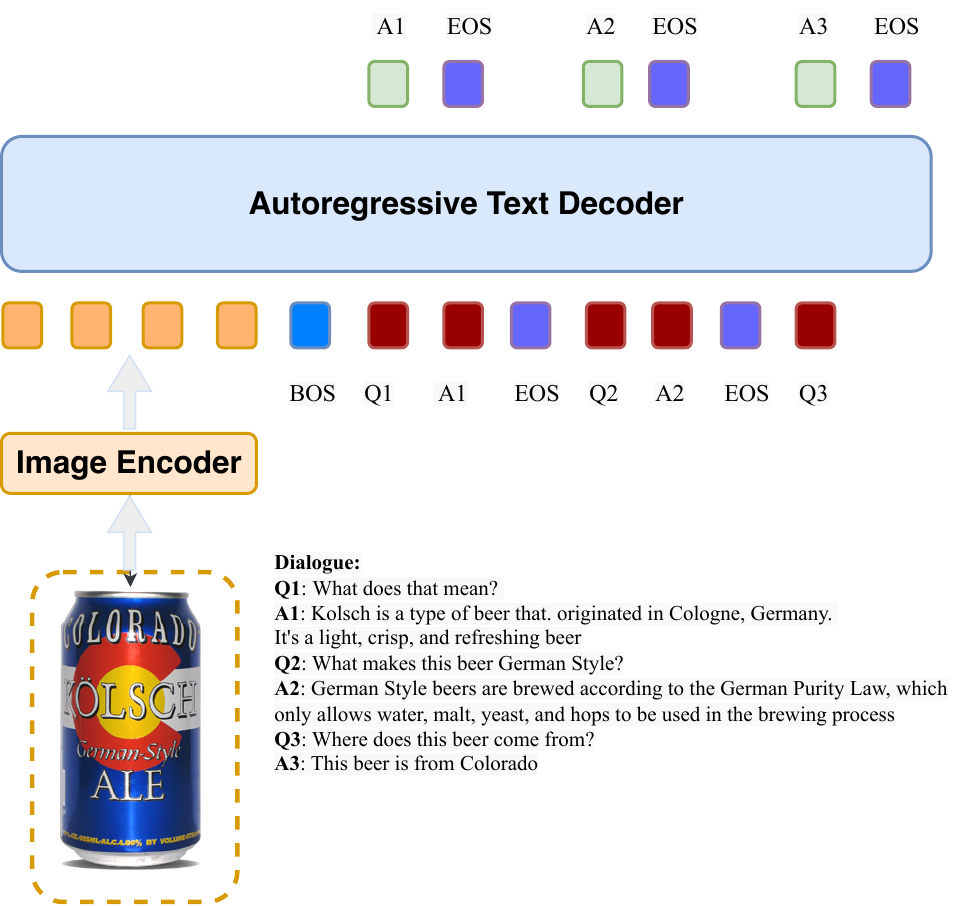}
  \caption{Finetuning GIT on InfoVisDial Dataset.}\label{fig:model}
\endminipage\hfill
\end{figure*}

\vspace{2mm}
\noindent\textbf{Dataset curation.} Although human-annotated datasets are expensive and subjective~\cite{schwartz-etal-2017-effect}, crowdsourcing remains the most popular method for creating new datasets. Prior work on automatic data curation mainly focuses on extraction methods and pattern matching~\cite{li2021guided}.
With the ever-growing capability of large pre-trained language models, some recent works study automatic dataset curation methods by finetuning LMs on existing datasets~\cite{anaby2020not,yang-etal-2020-generative}.
In this study, we provide a solution for automatic data curation via bridging multimodal and language foundation models.

%% file: approach.tex
% \begin{figure}[t]
% \begin{center}
%   \centerline{\includegraphics[width=0.8\linewidth]{latex/figure/prompt.pdf}}
% \end{center}
% \vspace{-0.3in}
%     \caption{Example of few-shot prompt.}
% \label{fig:prompts}
% \end{figure}

% \begin{figure}[t]
% \begin{center}
%   \centerline{\includegraphics[width=0.8\linewidth]{latex/figure/model.pdf}}
% \end{center}
% \vspace{-0.2in}
%     \caption{Finetuning GIT on InfoVisDial Dataset.}
% \label{fig:model}
% \end{figure}

% \begin{figure*}[t]
% \minipage{0.45\textwidth}
%   \includegraphics[width=\linewidth]{latex/figure/prompt.pdf}
%   \caption{Example of few-shot prompt.}\label{fig:filter_anno}
% \endminipage\hfill
% \minipage{0.45\textwidth}
%   \includegraphics[width=\linewidth]{latex/figure/model.pdf}
%   \caption{Finetuning GIT on InfoVisDial Dataset.}\label{fig:question_length}
% \endminipage\hfill
% \end{figure*}

\section{Methodology}

In Figure~\ref{fig:framework},
we propose a new framework to automatically curate the informative visual dialogue dataset by bridging multimodal and language foundation models.

%%%%%%%%%%%%%%%%%%%%%%%%%%%%%%%%%%%%%%%%
\subsection{Problem Formulation}
% This work aims to build flexible, open-ended and informative visual dialogue tasks that could enhance the chatting ability of vision and language models and benefit existing visual dialogue tasks.
Given a question $Q_k$ grounded on an image $I$ at the $k$-th turn and the previous dialogue history $H_t = (Q_1,A_1),(Q_2,A_2),...,(Q_{k-1},A_{k-1})$, our task aims to generate the target answer.

\subsection{LLM to InfoVisDial: Knowledge Distillation}
We propose to address the knowledge-lacking problem in existing visual dialogues by harnessing the in-context few-shot learning capability of large language models (LLM) like GPT-3. 
% Instead of directly deploying GPT-3 for downstream tasks, we leverage GPT-3 to achieve a more cost-effective and efficient training of vision-language models. 

\vspace{2mm}
\noindent\textbf{Prompting methods.}
The goal is to design a prompt appropriately such that GPT-3 can generate a dialogue.
Specifically, our designed prompts
% The few-shot prompt 
consist of three parts: instruction, few-shot context and image textual descriptions. The instruction is a concise description of the generation task. \eg, ``Generate a longer conversation about an image based on the context that describes the image content. Casey asks questions about the image content, and Riley answers them based on the context.'', which has shown its effectiveness in zero-shot and few-shot settings~\cite{brown2020language,wei2021finetuned}. The few-shot context contains in-context examples shown in Figure~\ref{fig:prompts}. These dialogues work as a shared template for all tasks. We design two kinds of image descriptions: ($i$) ground-truth captions from TextCap dataset; ($ii$) captions generated by pre-trained vision-language models (\emph{e.g.}, GIT~\cite{wang2022git}).
In order to generate longer conversations, we 
% further 
inject this (add the word of \textit{longer}) explicit intention into instructions and few-shot context. 
% We first propose to insert "longer" into the instruction. And then we adopted few-shot contexts instead of zero-shot contexts. 
While constructing the in-context examples, we found that using 
fewer 
but longer dialogues are more effective than using more but shorter dialogues. % when dealing with the limitation of the length of in-context examples. 

\noindent\textbf{Human in the loop.} Inspired by \cite{wang-etal-2021-want-reduce,yoo-etal-2021-gpt3mix-leveraging}, we adopt a human-in-the-loop annotation strategy to further study the quality of GPT-3 generated dialogue. Existing visual dialogue datasets~\cite{das2017visual} adopt fully human-annotated methods to let two annotators write questions and answers respectively, which is super expensive and labor-intensive. Instead, we design a simple and easy annotation method to obtain a good trade-off between annotation cost and dialogue quality. After generating a big dialogue corpus, we involve humans in the fact checking process. 
We ask 1 annotators to rate each round of dialogue given an image and its complete dialogue as context, with options including ``1: the question is relevant to the image and answer is correct'', ``2: the question is relevant to the image and I'm not sure answer is correct'', ``3: the question is relevant to the image and answer is not correct'' and ``4: the question is not relevant to the image''. The details of human annotation is provided in the Appendix. The filtering strategies will be introduced in Section~\ref{sec:dataset}.
By mixing GPT-3 and the human-in-the-hoop method, we could collect a benchmark dataset for informative visual dialogue at scale.

\begin{table}[t]
\centering
\scalebox{0.9}{
\begin{tabular}{cccccccccc}
\toprule
 Filtering strategy & Train & Val & Overall \\ \midrule
Keep all &346676 & 51000& 397675\\
keep the longest &69335 &10200 &  79535\\
Keep rate1 &42023& 5601 & 47524\\
\bottomrule
\end{tabular}
}
  \caption{There are three filtering strategies and their corresponding dataset size. InfoVisDial adopts ``keep the longest'' in the training set and ``keep rate1'' in the evaluation set.  }
\label{tab:filter_four}
\end{table}

\begin{table}[t]
\centering
\scalebox{0.9}{
\begin{tabular}{cccccccccc}
\toprule
  & Train & Val & Overall \\ \midrule
questions &69335 &5601 & 74936\\
dialogues &21953 &2086 &  24039\\
tokens/question &5.9& 6.1 &5.9\\
tokens/answer &8.9 &8.8 &8.9\\
questions/dialogue &3.0 &3.1 &3.0\\
\bottomrule
\end{tabular}
}
  \caption{Statistics summarizing the InfoVisDial dataset.}
\label{tab:dataset stats}
\end{table}

\begin{figure*}[t]
\minipage{0.32\textwidth}
  \includegraphics[width=\linewidth]{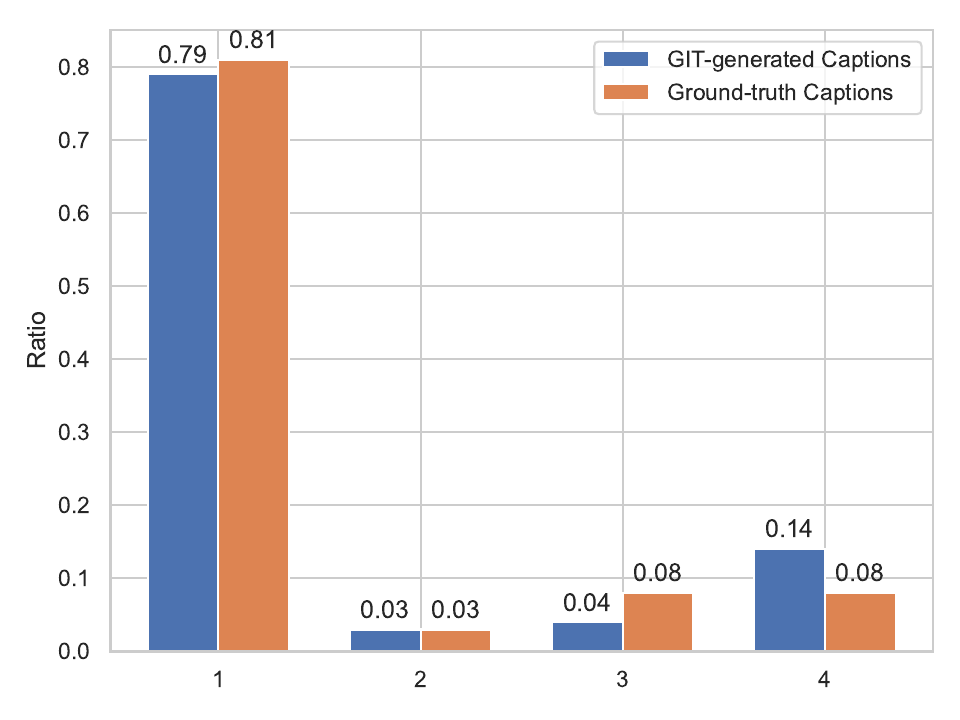}
  \caption{Human evaluation of the dialogue quality in InfoVisDial.}\label{fig:filter_anno}
\endminipage\hfill
\minipage{0.32\textwidth}
  \includegraphics[width=\linewidth]{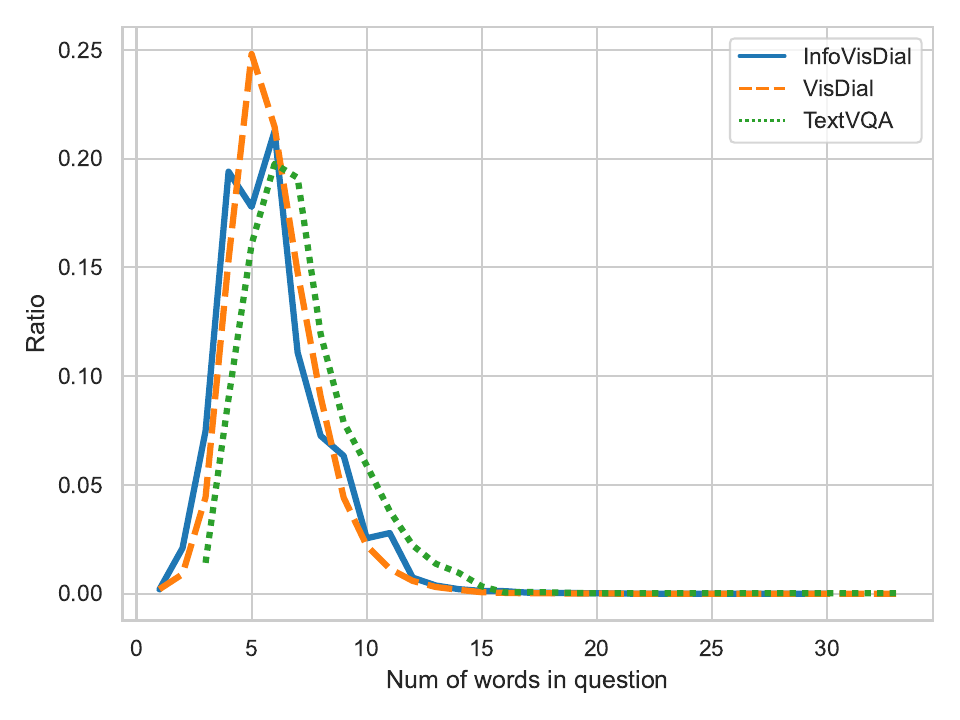}
  \caption{Distribution of question length.}\label{fig:question_length}
\endminipage\hfill
\minipage{0.32\textwidth}%
  \includegraphics[width=\linewidth]{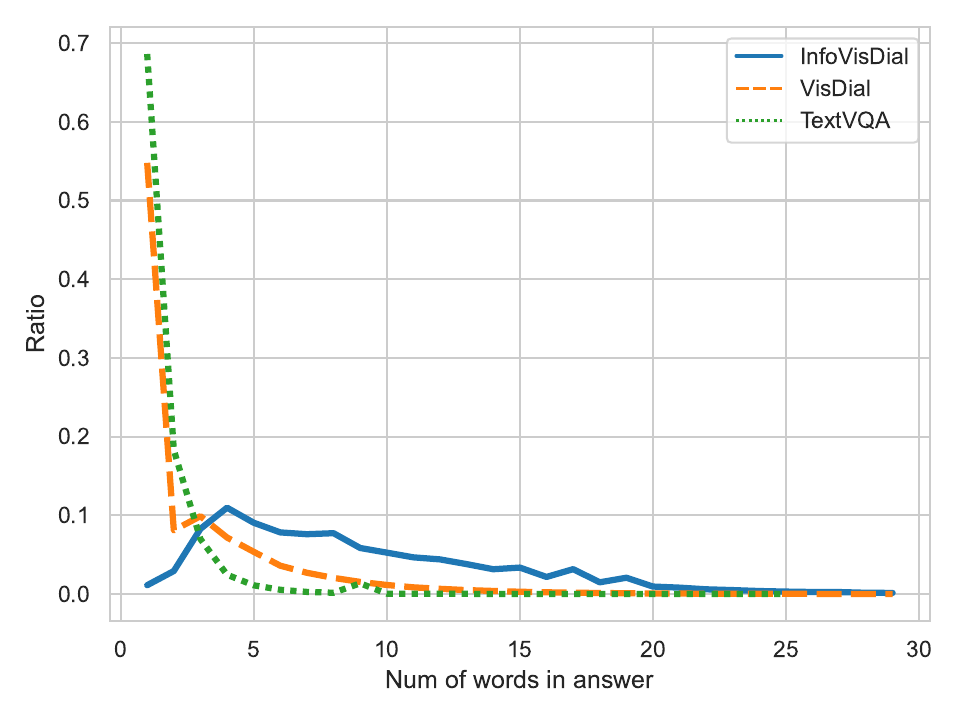}
  \caption{Distribution of answer length.}\label{fig:answer_length}
\endminipage
\end{figure*}

\begin{figure*}[t]
\minipage{0.32\textwidth}
  \includegraphics[width=\linewidth]{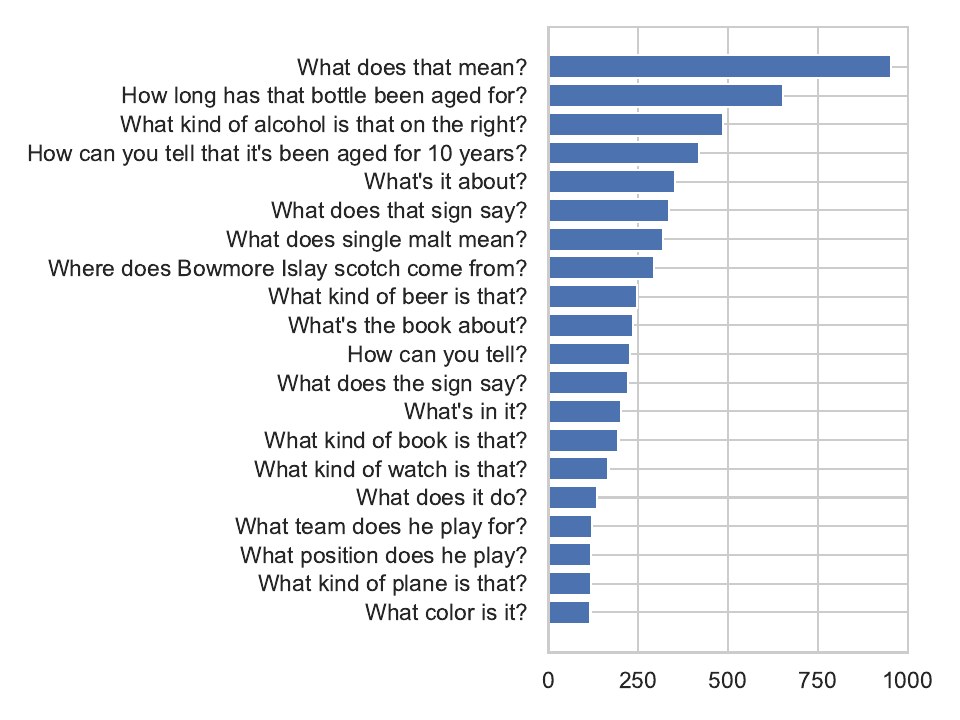}
  \caption{Top 20 most occurring questions in InfoVisDial.}\label{fig:top_question}
\endminipage\hfill
\minipage{0.32\textwidth}
  \includegraphics[width=\linewidth]{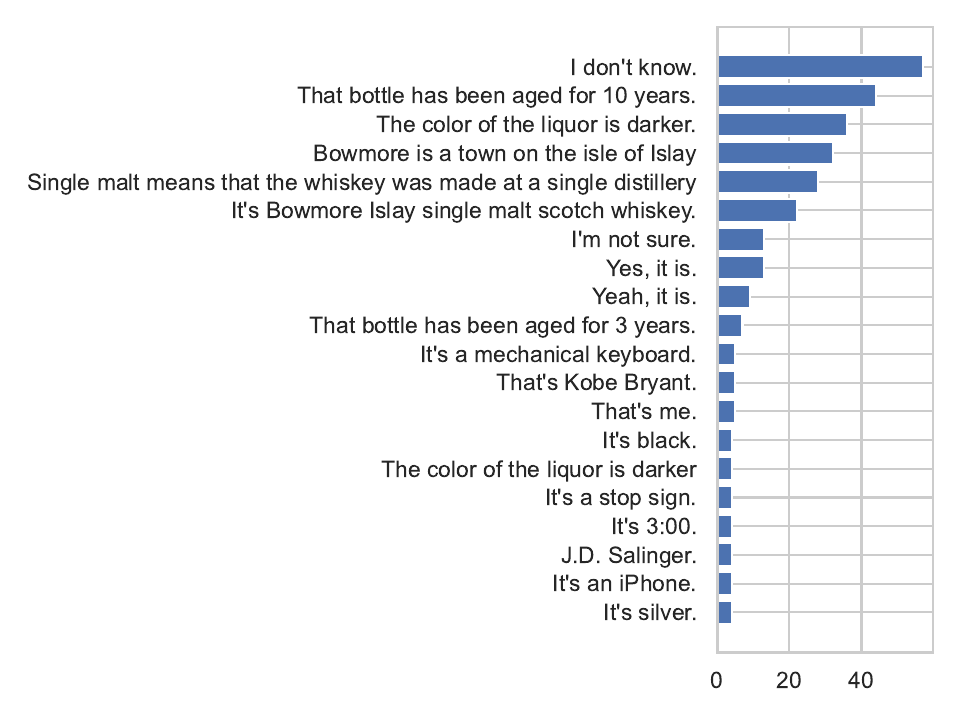}
  \caption{Top 20 most occurring answers in InfoVisDial.}\label{fig:top_answer}
\endminipage\hfill
\minipage{0.32\textwidth}%
  \includegraphics[width=\linewidth]{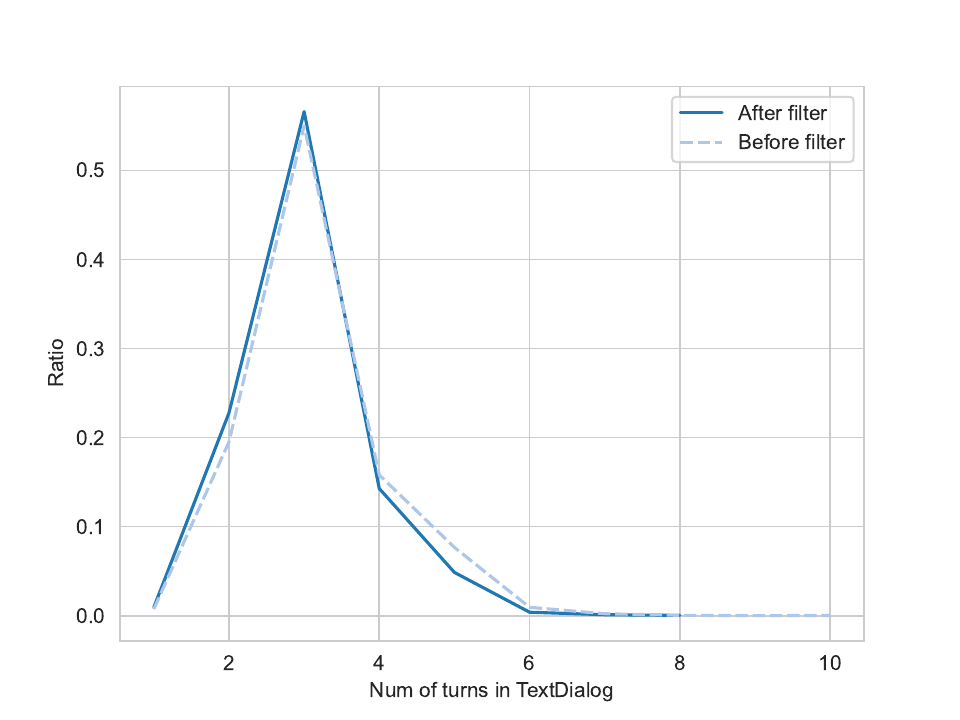}
  \caption{Distribution of dialogue length.}\label{fig:dialog_length}
\endminipage
\end{figure*}

\begin{figure}[t]
\begin{center}
  \centerline{\includegraphics[width=\linewidth]{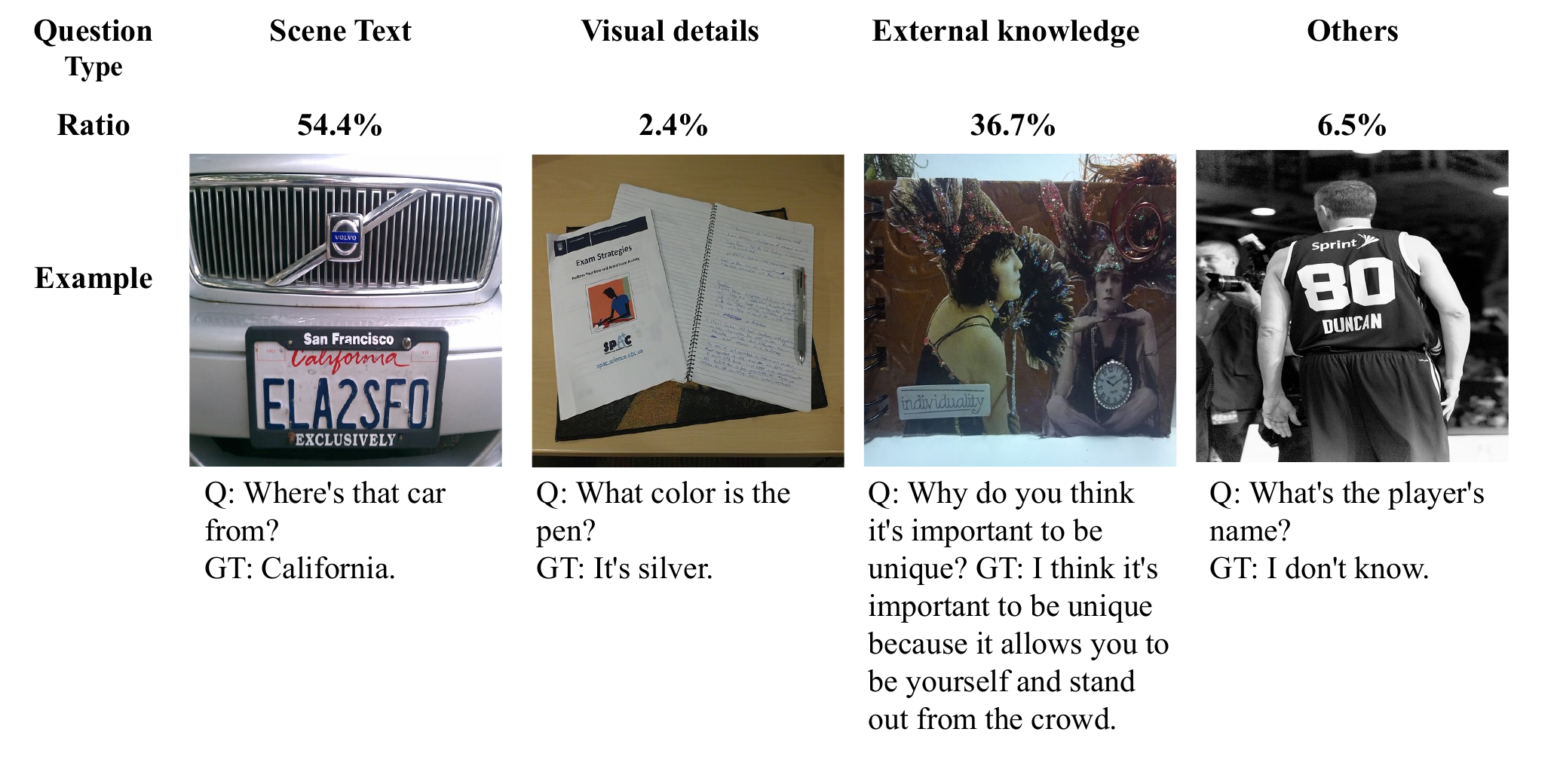}}
\end{center}
\vspace{-0.3in}
    \caption{Examples of four types of questions.}
\label{fig:question_type}
\end{figure}

\subsection{InfoVisDial to VLM: Knowledge Transferring}
% \zyang{to replace section name with model name.}

The final step of our framework is
to transfer the dialogue ability to vision-language models by finetuning GIT~\cite{wang2022git} on InfoVisDial, as shown in Figure \ref{fig:model}.
% Following previous work on vision-language pre-training~\cite{dosovitskiy2020image, yuan2021florence}, we adopt a Swin-like vision transformer as image encoder, thus getting rid of the dependency on the object detector which is widely used in almost all visual dialogue models~\cite{das2017visual,lu2017best,kottur2018visual,niu2019recursive,gan2019multi,guo2019image,murahari2020large,nguyen2020efficient,zheng2019reasoning,chen2022utc,wang2020vd,chen2020counterfactual,agarwal2020history}. 
An image encoder is used to extract the image features and a text decoder transformer
is 
to predict the text tokens with the language modeling task.
% The text decoder is a multi-modal transformer which consists of multiple transformer blocks. 
The dialogue is converted into one special \textit{image caption}
by concatenating all the question answer pairs separated by the \texttt{[EOS]} token.
In the training, the loss is imposed only on the answer part;
while in the inference, the model is able to predict the answer based on the 
dialogue history and current question.
If the dialogue has only one question-answer pair, the fine-tuning strategy is 
degenerated to the case of how GIT is adapted to the VQA task as presented in~\cite{wang2022git}.
% The text input is as follows:
% \begin{equation*}
%     % H = \left\{[BOS]Q_1A_1[SEP]Q_2A_2[SEP]Q_{k-1}A_{k-1}Q_{k}[MASK]\right\}
%     \left\{[BOS]Q_1A_1[SEP]Q_2A_2[SEP]Q_{k-1}A_{k-1}Q_{k}[MASK]\right\}
% \end{equation*}
% where the answer tokens $A_k$ are fully masked. 
% During inference, the answer is totally removed. We concatenate the dialogue history and answer as text input. Text input starts with the token $[BOS]$. Then we concatenate the image features and text embeddings as the input to the text decoder. Text token is decoded in an auto-regressive way， which means it only depends on the preceding tokens and all image tokens. The decoding process is stopped when the token $[EOS]$ is emitted or reaches the maximum step.

%% file: dataset.tex
\section{InfoVisDial Dataset}\label{sec:dataset}
%%%%%%%%%%%%%%%%%%%%%%%%%%%%%%%%%%%%%%%%%%%%%%%%%%%%%%%%%%%%%%%%%

In order to study the task of informative visual dialogue, we collect a new dataset named InfoVisDial. In this section, we start with dataset collection pipleline. We then provide statistics and an analysis of the dataset. Finally, we formulate two tasks based on the dataset.

\subsection{Dataset collection}
We use TextVQA dataset~\cite{singh2019towards} as the source of our images. Most images in TextVQA dataset contain informative text, which is closer to our real life.  We collect InfoVisDial's training and validation set from TextVQA's training while the evaluation set is collected from TextVQA's validation set. 

To collect dialogue for each image, we generate dialogue according to our proposed framework. Given an image and its textual representation, we repeat the prompting GPT-3 process five times and get five different dialogues. We put these dialogue candidates into the dialogue corpus. Then we involve humans in the loop to perform fact-checking. We propose three different filtering strategies. Keep all means no filtering. Keep the longest means only keeping the one with the longest dialogue length. Keep one means only keeping the dialogues with all rate1 in all turns. Not surprisingly, keep rate1 has the smallest dataset size. More details about filtering strategies is shown in Table \ref{tab:filter_four}.

Figure~\ref{fig:filter_anno} shows 78.88\% of QA pairs are labeled as rate1 which means the question is reasonable and the answer is correct. 13.88\% of QA pairs are labeled as the question is not relevant to the image. Not surprisingly, dialogues with GT captions have better quality than those with GIT-generated captions.

We conduct diversity analysis to see whether ``keep rate1'' filtering strategy hurts diversity. We surprisingly observe proportionally more diverse examples. One possible reason is that if GPT-3 prefers to generate some type of question and answer, these would be highly recognizable across human filtering, and removing the would increase both quality and diversity.

\subsection{Statistics and Analysis}

\noindent\textbf{Analyzing questions.}
We first analyze the diversity of the questions in the dataset. InfoVisDial contains 69335 questions and 56.2\% questions are unique. Figure~\ref{fig:question_length} shows the distribution of question length along with the statistics for the VisDial and TextVQA datasets for reference. The average question length in InfoVisDial is 5.93 words which are similar to VisDial(5.96) but less than TextVQA (7.18). Unsurprisingly, the average length of questions from the dialogue dataset is usually short than the length of questions from QA tasks. We also note that the minimum length of questions from InfoVisDial is 1 which is the same as VisDial, while the minimum length of questions from TextVQA is 3.   Some questions whose length is 1 like ``Why'', ``Who'', ``How'' and ``So'' are prevalent in conversations. We ask annotators to label the type of questions. The details of human evaluation will be included in the supplementary materials. Figure~\ref{fig:question_type} illustrates the annotation result. 54.4\% questions are about scene text, 36.7\% questions are about external knowledge and 2.4\% questions are about visual details. Figure~\ref{fig:top_question} shows top 20 most occurring questions in the dataset with their count.  We could see We find GPT-3 is more likely to generate questions that appear in the prompts. Even after human filtering, the distribution still skewed to the prompt questions. We find the question type becomes more diverse after filtering all the prompt questions. We will leave the task of building a more balanced dataset as future work. We include more question analysis figures in supplementary materials.  
% Figure~\ref{fig:question_cloud} illustrates wordcloud of questions

\noindent\textbf{Analyzing answers.} Figure~\ref{fig:answer_length} shows the distribution of answer lengths. InfoVisDial contains 69335 answers and 87.3\% answers are unique. The percentage is quite high compared to TextVQA (49.2\%) and VisDial(27.37\%). The average answer length in InfoVisDial is 8.9 which is much longer than VisDial(2.9). Figure~\ref{fig:top_answer} shows the top 20 most occurring answers in the dataset with
their count. We find answers in InfoVisDial are longer, more informative and descriptive. Some answers of InfoVisDial are extremely long because they include external knowledge. \eg, when asked about the specialty of Rollei, the answer will be ``The Rollei 35 is a compact 35mm film camera that was first introduced in 1966. It's special because it was one of the first 35mm cameras to be designed as a compact camera''. We also find that answers appear on the prompts have a higher possibility to be generated and will apply the same filtering approach as questions in the future. We include more answer analysis figures in supplementary materials.  
% Figure~\ref{fig:answer_cloud} illustrates wordcloud of answers.

\noindent\textbf{Analyzing dialogue.} Figure~\ref{fig:dialog_length} shows the average turns of InfoVisDial is 3.0. We could see the majority of dialogues have no less than 3 turns. We find that 12\% of dialogues require reading the dialogue history to resolve coreferences by human annotation. We compare the length of dialogue before and after filtering. We find they follow similar distribution in general. But filtering does slightly decrease the portion of dialogues whose lengths are more than 4. Finally, we examine the properties of the question in various turn positions in the dialogue. The frequency of yes/no questions increases significantly as the dialogue progress; The percentage of hard questions also increases.

%% file: exp.tex
\begin{table*}[t]
\centering
\scalebox{1}{
\begin{tabular}{lccccccccc}
\toprule
Method & F1&CIDEr& B\textsubscript{1}& B\textsubscript{2}&B\textsubscript{3}&B\textsubscript{4}&METEOR&R\textsubscript{L}&SPICE \\ \midrule
Fine-tuned GIT\textsubscript{B}&26.0& 95.9 & 27.4&21.2 & 17.2 &14.7 & 12.7 & 32.4 &12.1 \\
 Fine-tuned GIT\textsubscript{L} &32.9&143.3& 32.7&26.1&21.6&18.4&16.1&37.8&18.5 \\
Fine-tuned GIT &35.9 &170.9&35.5&28.5&23.7&20.2&17.9&40.3&21.9 \\
\bottomrule
\end{tabular}}
  \caption{We follow caption metrics and word-level F1~\cite{choi-etal-2018-quac} to evaluate the performance of our model on InfoVisDial dataset on the single round VQA training setting.}
\label{tab:singlevqa}
\end{table*}

\begin{table*}[t]
\centering
\scalebox{1}{
\begin{tabular}{lccccccccc}
\toprule
Method & F1&CIDEr& B\textsubscript{1}& B\textsubscript{2}&B\textsubscript{3}&B\textsubscript{4}&METEOR&R\textsubscript{L}&SPICE \\ \midrule
Fine-tuned GIT\textsubscript{B} &11.9 &54.5& 13.2&10.0 &7.8 &6.1&7.9 &23 &6.8\\
Fine-tuned GIT\textsubscript{L} &16.2 &93.6& 19.7&15.4&12.3&10.15&11.8&30.1 &13.8 \\
Fine-tuned GIT &18.8 &134.3&22.5&17.9&14.5&11.9&13.4&33.7&17.1 \\
\bottomrule
\end{tabular}}
  \caption{We follow caption metrics and word-level F1~\cite{choi-etal-2018-quac} to evaluate the performance of our model on InfoVisDial dataset on the multi round VQA training setting.}
\label{tab:mul_vqa}
\end{table*}

\begin{figure}[t]
\begin{center}
  \centerline{\includegraphics[width=\linewidth]{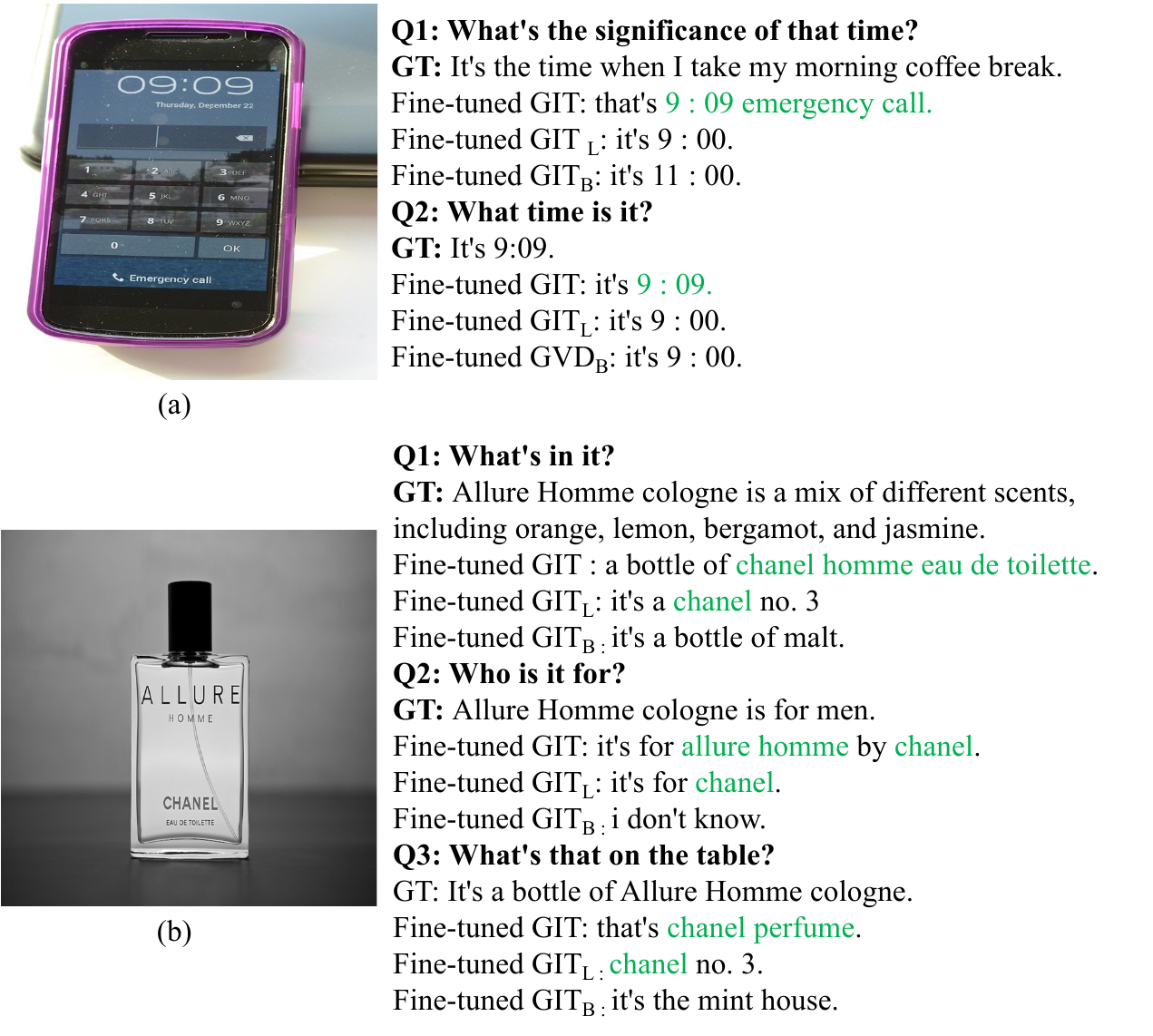}}
\end{center}
\vspace{-0.3in}
    \caption{Results comparison between different size of fine-tuned GIT models.}
\label{fig:model_comp}
\vspace{-3mm}
\end{figure}

\section{Experiments}
%%%%%%%%%%%%%%%%%%%%%%%%%%%%%%%%%%%%%%%%%%%%%%%%%%%%%%%%%%%%%%%%%
We conduct experiments on our newly proposed task and dataset InfoVisDial to show that our model Fine-tuned GIT achieves great performance on dialogue ability. Further ablation studies and a qualitative analysis provide more insights into our approach.
% We also perform data augmentation experiments on
% visdial task and show that our method hugely boosts the performance of visdial task. 

\subsection{Experiment setup}

\noindent\textbf{Dataset.} To evaluate our proposed model, we conduct extensive experiments on our proposed dataset InfoVisDial. InfoVisDial dataset has 21953 and 2086 images for training and evaluation, respectively. Each image is associated with a open-ended and free-text form dialogue. 

\noindent\textbf{Evaluation metrics.} We propose to use caption and word accuracy~\cite{choi-etal-2018-quac} metrics to evaluate the text quality and content accuracy, respectively. Inspired by~\cite{yeh2021comprehensive}, we conduct a user study to show that our metrics align with human judgments. 

% \noindent\textbf{VisDial dataset} We evaluate our proposed approach on the VisDial v1.0 dataset. It has 123287, 2064, and 8000 images for training, validation and testing, respectively. Each image is associated with a caption sentence and 10 question-answer pairs. For each round of question-answer pairs, 100 answer candidates are given. The validation split and part of the train split (2,000 images) are provided with dense annotations (\ie, relevance scores) for all candidate answers. We follow their standard evaluation protocol for evaluating our model. 

\begin{figure}[t]
\begin{center}
  \centerline{\includegraphics[width=\linewidth]{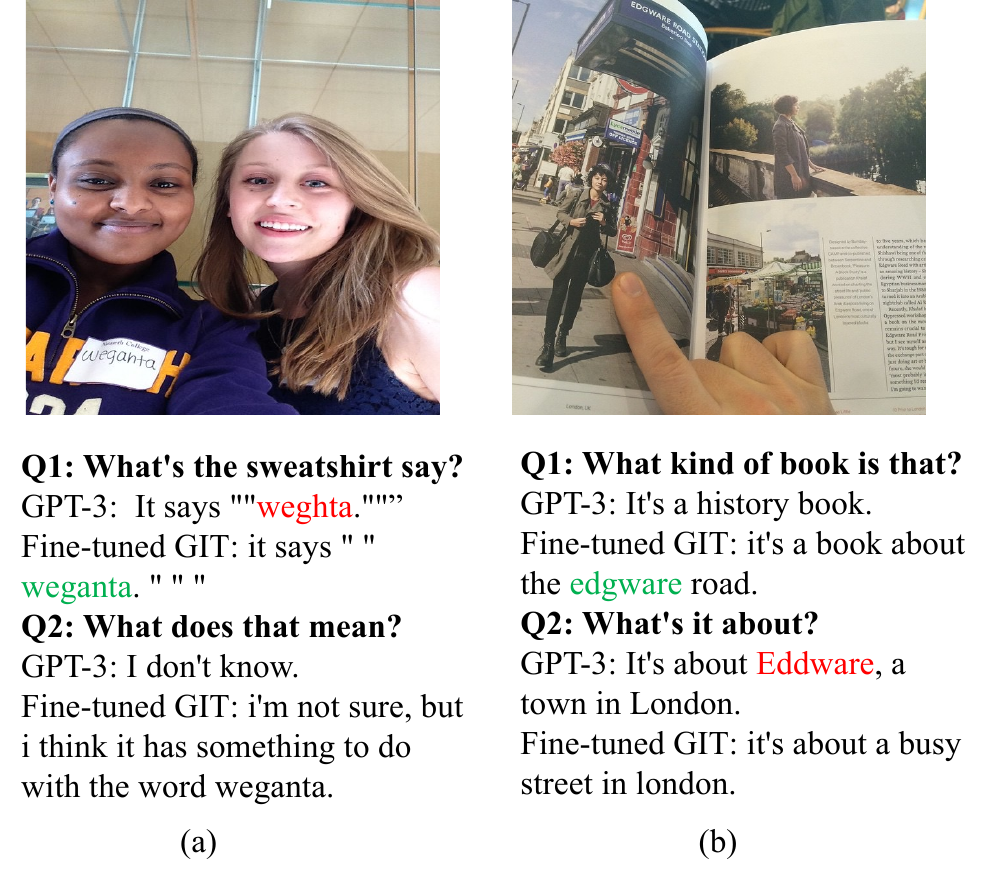}}
\end{center}
\vspace{-0.3in}
    \caption{Comparison the results between the GPT-3 (without human cleaning) and our fine-tuned GIT. We could see our model fine-tuned GIT has better visual understanding ability in terms of scene text.}
\label{fig:gpt3_comp}
%\vspace{-3mm}
\end{figure}

\noindent\textbf{Implementation Details.} In this study, we use GPT-3 \textit{text-davinci-002} API to perform the prompting. Our model baseline is Fine-tuned GIT~\cite{wang2022git}. GIT-large and GIT-small are trained on two smaller pre-training dataset. GIT adopts coswin as image encoder~\cite{yuan2021florence}. GIT-large uses CLIP/ViT-L/14~\cite{radford2021learning} as image encoder. GIT-small uses CLIP/ViT-B/16~\cite{radford2021learning} as image encoder. The number of training epoch is 20 and the learning rate is 1e-5. The input image size is 576 for GIT, 480 for GIT-large and 560 for GIT-small. The batch size is 16 and parameter optimizer is AdamW~\cite{loshchilov2017decoupled}.

\subsection{Experiments on InfoVisDial}
We consider the following two training settings: single round VQA and multi round VQA. Like existing dialogue tasks, multi round VQA need to predict the answer based on dialogue history. Meanwhile, we also introduce single round VQA which has simple settings and could be easily used as benchmark for other tasks.

\vspace{2mm}
\noindent\textbf{Single-round VQA.} We compare the results from various size of Fine-tuned GIT models shown in Table~\ref{tab:singlevqa}. GIT has much better performance both on content accuracy and text quality compared to large and base models. Figure~\ref{fig:model_comp} shows some qualitative comparison. We could see Fine-tuned GIT has great ability to read scene text in the image. In Figure~\ref{fig:model_comp} (a), when asked ``what's the significance of that time'', Fine-tuned GIT not only could read and understand that ``9:09'' is time, but also connects ``emergency call'' with ``significance''. Although it doesn't answer the question in a perfect way due to lack of background knowledge, it gives a reasonable response. In Figure~\ref{fig:model_comp} (b), Fine-tuned GIT also gives more relevant answers than the large and base model. When asked ``What's that on the table?'', Fine-tuned GIT's answer ``that's chanel perfume'' is quite reasonable. We observe that Fine-tuned GIT-base easily answers ``I don't know'' or give totally irrelevant answers. 

\vspace{2mm}
\noindent\textbf{Metrics evaluation and estimating human performance.} To validate the effectiveness of our metrics, we conduct a user study to ask annotator rate the turn-level dialogue answer on a likert scale. We find our proposed metrics are significantly correlate with human judgments.  We also estimate an upper bound on model performance given our metrics, we assess the answers that annotated as ``5''. Then we calculate the human F1 based on these answers. Note that this is a lower bound on human performance. Details about the user study will be introduced in the Appendix.
% \begin{table*}[t]
% \centering
% \scalebox{1}{
% \begin{tabular}{ccccccccc}
% \toprule
% Method & Val Acc & Test Acc \\ \midrule
% previous SOTA \\
% GIT_B  & & \\
% GIT_L  & & \\
% GIT  & & \\
%  \textbf{Fine-tuned GIT}_B & & \\
%  \textbf{Fine-tuned GIT}_L & & \\
%  \textbf{Fine-tuned GIT} & & \\
% \bottomrule
% \end{tabular}
% }
%   \caption{We further test our \textbf{Fine-tuned GIT} on TextVQA dataset. Our  \textbf{Fine-tuned GIT} achieves comparable performance on TextVQA dataset, which shows our \textbf{Fine-tuned GIT} could read the visual content and learn knowledge from dialogues effectively.   }
% \label{tab:TextVQA}
% \end{table*}

% \begin{table}[t]
% \centering
% \scalebox{1}{
% \begin{tabular}{ccccccccc}
% \toprule
% CIDEr/F1 \\ \midrule
% Model & 1x & 5x& 10x& 20x& 30x \\ \midrule
% \textbf{Fine-tuned GIT}_B & &  &  &  &  \\
% \textbf{Fine-tuned GIT}_L & & & &  & \\
% \textbf{Fine-tuned GIT} & & & &  &\\
% \bottomrule
% \end{tabular}
% }
%   \caption{Result of our model in the high-data regime. We report CIDEr of the different size of models on the InforVisDial val split.
%   Apart from the InforVisDial data, we utilize CC3M as extra training data (1, 5, 10, 20, and 30 times of InforVisDial train split.)}
% \label{tab:TextVQA}
% \end{table}

\vspace{2mm}
\noindent\textbf{Does fine-tuned GIT overcome the visual information bottleneck of GPT-3?} One motivation of proposing fine-tuned GIT instead of just prompting GPT-3 is that GPT-3 cannot fully understand visual information through textual description. Therefore, we validate if fine-tuned GIT overcomes this limitation. We compare the dialogue generated by fine-tuned GIT and GPT-3 (without filtering) respectively based on the GIT-generated captions. In Figure~\ref{fig:gpt3_comp} (a), GPT-3 is misled by the incorrect textual description ``weghta'' while fine-tuned GIT could recognize the scene text correctly. We also observe that fine-tuned GIT generates more human-like answers when facing hard questions. Instead of just answering ``I don't know'', fine-tuned GIT makes a reasonable guess ``but i think it has something to do with the word weganta''. In Figure~\ref{fig:gpt3_comp} (b), fine-tuned GIT provides correct answer about scene text and external knowledge. 

% In knowledge distillation, the student model is generally expected to lose some performance~\cite{hinton2015distilling, kim2016sequence} compared to its teacher. 
\vspace{2mm}
\noindent\textbf{Multi-round VQA.} Fine-tuned GIT demonstrates its effectiveness both on content accuracy and text quality
compared to fine-tuned GIT-large and GIT-base models. We observe a performance decrease in multi-round training setting in Table~\ref{tab:mul_vqa}. There are several possible reasons: ($i$) The current model may not be good at modeling long sequences, since they are pre-trained on caption datasets whose length are much smaller than dialogue history. ($ii$) The answer in each turn contains a large amount of external knowledge. Simple concatenation would not catch the clues in dialogue history. We come up with two solutions and leave them to future work:  ($i$) strengthening the model's long sequence modeling ability; ($ii$) data augmentation. We are also building a larger-scale informative visual dialogue dataset based on CC3M~\cite{changpinyo2021cc12m}, and will also add them into the training set.

\vspace{1mm}
\noindent\textbf{Error analysis.} To gain insights into the model’s errors, we sample 50 evaluation questions with predicted F1 scores below 0.10 from the fine-tuned GIT model trained under single VQA training setting. We remove cases in which the predicted answers are actually correct. We summarize the common errors in the Appendix.

\noindent\textbf{Prompt tuning for generative dialogue.}
Although we find one-shot prompt will generate some dialogue which is similar to the one-shot example, it will be easily recognized and filtered by annotators.
One-shot prompt could generate similar quality of dialogue than few-shot prompts. More details about different prompt designs are provided in the Appendix.

% \begin{table}[t]
% \centering
% \scalebox{1}{
% \begin{tabular}{cccccccccc}
% \toprule
% Prompting strategy & Train & Val & Overall \\ \midrule
% n=0 &346676 & 51000& 397675\\
% n=1&69335 &10200 &  79535\\
% n=2 &42023& 5601 & 47524\\
% n=3 &42023& 5601 & 47524\\
% \bottomrule
% \end{tabular}
% }
%   \caption{There are various prompting strategies and their generative dialogue analysis. n is the number of in-context examples}
% \label{tab:filtering}
% \end{table}

%% file: conclusion.tex
% \section{Limitation}

\section{Conclusion}

We have revisited the visual dialogue problem and presented a new dataset named InfoVisDial. InfoVisDial contains informative visual dialogues that require the model to read scene text and reason about relevant knowledge to generate. We automatically curate this dataset by proposing a new framework that synergistically combines the image-to-text GIT captioning model and the GPT-3 language model. Human analyses show that this scalable data collection framework leads to the desired informative dialogue dataset. InfoVisDial contains dialogues that are diverse and informative in topics, with free-form long answers. Furthermore, we propose a strong baseline by fine-tuning GIT on the InfoVisDial dataset\eat{ to transfer the informative dialogue ability}.
% Then We conduct an extensive experiments to show the effectiveness of fine-tuned GIT and also demonstrate the difficulty of the tasks in InfoVisDial. 
We hope our proposed dataset would provide a solid playground for the visual dialogue community.

% In this paper, we build a visual dialogue dataset, named InfoVisDial, which provides rich informative answers even with external knowledge related to the visual content. Different from existing datasets where the answer is compact and short, InfoVisDial contains long free-form answers with rich information in each round of dialog. For effective data collection, the key idea is to bridge the large-scale captioning model (GIT) and the language models (GPT-3). GIT can describe the image content even with scene text, while GPT-3 can generate informative dialogue based on the image description and appropriate prompting techniques. With such automatic pipeline, we can readily generate informative visual dialogue data at scale. Then, we ask human annotators to rate the generated dialogues to filter the low-quality conservations. Last, we propose a simple adaptation of the GIT model for the visual dialogue task and fine-tune the model on InfoVisDial. Hopefully, our work can motivate more effort on this direction.